%% file: main_arXiv.tex
\documentclass{article}

\usepackage{microtype}
\usepackage{booktabs} 

\usepackage[accepted]{icml2020}

\usepackage{url}
\usepackage{graphicx}
\usepackage{grffile}
\usepackage{hyperref}
\usepackage{subcaption}
\usepackage{amsmath, amssymb, bm, bbm}
\usepackage{physics}
\usepackage{mathrsfs}
\usepackage{mathtools}
\usepackage[scr=boondoxo]{mathalfa}
\usepackage{cleveref}
\usepackage{wrapfig}
\usepackage{color}
\usepackage{lipsum}
\usepackage{multirow}
\usepackage[table,svgnames]{xcolor}
\usepackage{here}
\usepackage{enumitem}  

\usepackage{amsthm}
\theoremstyle{definition}

\newtheorem{theorem}{Theorem}
\newtheorem*{theorem*}{Theorem}

\newtheorem*{definition*}{Definition}

\newtheorem*{lemma*}{Lemma}

\newtheorem*{prop*}{Proposition}

\DeclareMathOperator*{\argmax}{arg\,max}

\usepackage{comment}

\begin{document}

\twocolumn[
\icmltitle{Meta Learning as Bayes Risk Minimization}

\icmlsetsymbol{equal}{*}

\begin{icmlauthorlist}
\icmlauthor{Shin-ichi Maeda}{pfn}
\icmlauthor{Toshiki Nakanishi}{pfn}
\icmlauthor{Masanori Koyama}{pfn}
\end{icmlauthorlist}

\icmlaffiliation{pfn}{Preferred Networks, Inc., Japan}
\icmlcorrespondingauthor{Shin-ichi Maeda}{ichi@preferred.jp}


\vskip 0.3in
]

\printAffiliationsAndNotice{} 

\begin{abstract}
Meta-Learning is a family of methods that use a set of interrelated tasks to learn a model that can quickly learn a new query task from a possibly small contextual dataset.  
In this study, we use a probabilistic framework to formalize what it means for two tasks to be related and reframe the meta-learning problem into the problem of Bayesian risk minimization (BRM).
In our formulation, the BRM optimal solution is given by the predictive distribution computed from the posterior distribution of the task-specific latent variable conditioned on the contextual dataset, and this justifies the philosophy of Neural Process.
However,  the posterior distribution in Neural Process violates the way the posterior distribution changes with the contextual dataset.
To address this problem, we present a novel Gaussian approximation for the posterior distribution that generalizes the posterior of the linear Gaussian model.
Unlike that of the Neural Process, our approximation of the posterior distributions converges to the maximum likelihood estimate with the same rate as the true posterior distribution. 
We also demonstrate the competitiveness of our approach on benchmark datasets.

\end{abstract}
\section{Introduction}
\label{sec:intro}
Meta Learning is a family of method that efficiently solves new tasks by solving many interrelated tasks, and has succeeded in solving problems that were difficult to solve with conventional supervised learning methods \cite{Vilalta02Meta, Finn17MAML, Chen19CloserLookMeta}. 
To the author's best knowledge, however, there has not been any study to date that has clarified which meta-learning  method is optimal for which problem setting.

In this paper we use the theory of Bayes Risk Minimization (BRM) to provide the answer to this question when the stochastic input/output relation in each task is determined by the task-specific latent variable.
We show that, when we cast meta-learning problem as BRM, the optimal solution is given by the predictive distribution computed from the posterior distribution of the latent variable conditioned against the contextual dataset. 
This result justifies the use of the predictive distribution in many previous studies of meta learning, such as \citep{Edwards17NeuralStat, Gordon18BayesianMeta, Garnelo18CNP2}.
However, the optimality of the predictive distribution cannot be guaranteed if one uses an approximation of the posterior distribution that violates the way the posterior distribution changes with the contextual dataset, and this is unfortunately the case for most of the aforementioned works. 
For example, the variance of the posterior in these works do not converge to $0$ as we take the size of the contextual dataset to infinity. 
Therefore, in addition to our theoretical claim about the BRM, we propose a novel approximation of the posterior distribution.
By leveraging the properties of exponential distribution, we can construct a generalization of the linear Gaussian model that can satisfy all properties of the posterior distribution while maintaining high representation power.

While our approximation is built on a certain set of regularity assumptions, we can make some of these assumptions valid by grouping the members of contextual dataset into smaller subsets and appealing to Bernstein-von Mises theory. 
Bernstein-von Mises theory also assures that, if the number of observations for each task is $N$, the variance of our posterior distribution has order $O(1/N)$, which is same as the order of theoretically optimal Gaussian posterior.

The form of our approximation is closely related to that of the neural process (NP) \cite{Garnelo18CNP2}, but differs from NP in that it weighs each member of the contextual data by the uncertainty measure.  
Our design naturally encourages the predictor to preferentially use members with smaller uncertainty.
We will demonstrate the efficacy of our method on one-dimensional function approximation and the room rendering problem used in \citet{Eslami18GQN}. 
We summarize our key contributions below:

 \begin{enumerate}
     \item We show that the predictive distribution computed from the posterior distribution of the latent variable given the contextual dataset is the optimal solution of a BRM problem. This result justifies the philosophy of Neural Process. 
     \item We propose a novel exponential-family approximation of posterior distribution, and show that it converges to Maximum Likelihood Estimate (MLE) with the same rate as the true posterior distribution.
     \item We demonstrate that our novel approximation has enough representation power to produce competitive results in standard benchmark datasets.

 \end{enumerate}

\section{Problem formulation}
\subsection{Overview of the Meta Learning} 

Before we formalize our problem, we first review the concept of meta learning in general. 
Meta learning is a family of methods that aims to use the knowledge learned in one task to learn another.
This is a feat that cannot be achieved using classical supervised learning because the purpose of supervised learner is to exclusively learn the input-output relationship for the task of interest. 
Meta learner on the other hand, aims to learn "how to use" the (possibly small) \textit{contextual information} to learn th intput-output relationship for an arbitrary query task.
The contextual information of task $k$ is often assumed to be of form $\mathcal{D}_k =\{(x^{(k)}_1,y^{(k)}_1), \cdots, (x^{(k)}_{N_k},y^{(k)}_{N_k} )\}$, 
where $x^{(k)}_n$ denotes the $n$-th input of task $k$ and $y^{(k)}_n$ denotes its corresponding output\footnote{We omit the superscript $k$ if we do not need to specify the dependence on the task index.}. 
If the I/O relation on the domain $k$ is described by 
$y = f_k(x)$,  meta learning algorithms aim to learn how to approximate $f_k$ using the contextual information $\mathcal{D}_k$.


The celebrated MAML\cite{Finn17MAML} assumes that each $f_k(x)$ can be written as $f(x; h_k)$ with some task $k$ specific parameter $h_k$ , and uses an update rule  $U(\mathcal{D}_k; \theta)$ to approximate $h_k$.
More precisely, MAML uses $U(\mathcal{D}_k; \theta) = \theta - \epsilon \frac{\partial Loss(h; \mathcal{D}_k)}{\partial h} \mid _{h=\theta}$ as their approximation of $h_k$, 
where $Loss(h; \mathcal{D}_k)$ is the loss function to be minimized for each task $k$ and $\theta$ is the \textit{common initial parameter of $f$} that acts as the task-agnostic parameter. 
Extension of these algorithms even go further to learn the parameter space with task-specific energy-landscape \citep{Nichol18Reptile, Lee18Tnet, Park19MetaCurv, Flennerhag19WarpGrad}.

Meanwhile, the family of methods that includes neural processes (NP) interprets the approximation $h_k$ as a hidden variable in probabilistic model $p(y| x; h_k)$. 
They use encoder to describe the approximate the posterior distribution $p(h_k | D_k)$, and use decoder to approximate the forward model $p(y| x; h_k)$ 
\citep{Edwards17NeuralStat, Garnelo18CNP2,Kim19ANP,Louizos19FNP,Gordon19ConvolutionalCNP}.

From bird's eye point of view, we can say that almost all meta-learning methods developed to date use the same framework, with differences only in the way they approximate $h_k$ and the way they use it in their inference models.
While MAML-type methods deterministically approximate $h_k$ using $U(D_k, \theta)$, NP-type methods infer $h_k$ probabilistically and use encoder to approximate $p(h_k | D_k ; \theta)$.  

Now, the natural question will be  "Is one approach better than another in some situation? If so, when?"    
To our best knowledge, there has not been a study that investigated this question.
It turns out that, if the underlying model is stochastic and if the objective function is Bayes Risk, there is an answer to this problem. 
We elaborate this claim in the next section.

\subsection{Bayes Risk Minimization} 
In this section we formulate the meta learning problem as a case of Bayes Risk Minimization (BRM). 
In meta-learning, we assume that we are given a pool of datasets that corresponds to a set of tasks. 
In our Bayesian framework, we assume that each $\mathcal{D}_k$ is a set of iid samples from the conditional distribution $ p(x^{(k)}_n, y^{(k)}_n| h_k)$ parameterized by the task-$k$ specific latent variable $h_k \in \mathbb{R}^d$.
The size $N_k$ of $D_k$ may differ across tasks. 
By defining $p(h)$, we can also define a distribution on the set of tasks.
This way, the whole generation process of meta-learning dataset $\{D_k ; k =1,...,M\}$ can be described by the joint distribution $p(y, x, h)$. 


To make predictions on the query task $t$, we need to estimate $p( y^{(t)}_n | x^{(k)}_t, h_t)$.
The inconvenient fact here is that the functional form of $p( y^{(t)} | x^{(t)}, h_t)$ is not known in advance, let alone the value of the latent variable $h_t$. We also need to estimate $p( y^{(t)}_n | x^{(k)}_t, h_t)$ using $\mathcal{D}_{all} = \bigcup_{k=1}^K \mathcal{D}_k$
 and $\mathcal{D}_t$.
What is the form of the distribution constructable from $\mathcal{D}_{all} \uplus \mathcal{D}_t$ that can best approximate  $p( y^{(k)}_n | x^{(k)}_n, h_k)$? 
This question can be formulated in the form of BRM.

Let us use $q(\cdot ;x^{(t)}, \mathcal{D}_{all},\mathcal{D}_t)$ to denote an arbitrary distribution on the domain of $y$ that is constructed from $x^{(t)}$, $\mathcal{D}_t$ and $\mathcal{D}_{all}$. 
In order to answer the question above, we would like to look for $q$ that minimizes
\begin{align}
   E_{\prod_k p(h_k)}&\left[ E_{\{\prod_{k=1}^K p(\mathcal{D}_k|h_k)\}p(\mathcal{D}_t,x_{*}^{(t)}|h_t)}\right.  \nonumber \\
   &\left[ 
   KL[p(y_{*}^{(t)}| x_{*}^{(t)}, h_t)| q(y_*^{(t)};x_*^{(t)}, \mathcal{D}_t,  \mathcal{D}_{all} )]\right],  \label{eq:bayes_risk}
\end{align}
This minimization problem is a case of BRM problem.
Luckily, the optimal $q$ can be analytically solved
\cite{AITCHISON75BayesRisk}. The solution is in the form of a predictive distribution:
\begin{align}
&p(y^{(t)} | x^{(t)}, \mathcal{D}_t, \mathcal{D}_{all})  \nonumber \\
=& \int p(y^{(t)} | x^{(t)}, h_t) p(h_t |x^{(t)}, \mathcal{D}_t, \mathcal{D}_{all}) dh_t \nonumber \\
=&\int p(y^{(t)} | x^{(t)}, h_t) p(h_t |x^{(t)}, \mathcal{D}_t) dh_t.
\label{eq:bayes_risk_base}
\end{align}
Thus, when interpreted in the context of BRM, the task of meta-learning is to find the predictive distribution in Eq.\eqref{eq:bayes_risk_base}. 
To evaluate this integral, we need both $p(h_t |x^{(t)}, \mathcal{D}_t)$ and $p(y^{(t)} | x^{(t)}, h_t)$.
The former can be considered as a probabilistic encoder that maps $(x^{(t)}, \mathcal{D}_t)$ to $h_t$, and the latter can be considered as a decoder that probabilistically maps $(x^{(t)}, h_t)$  to $y^{(t)}$. 
Our job is now to learn this pair of encoder and decoder.



%
To make this learning problem tractable, meta learning often assumes some type of invariance relations to hold for  $p(x^{(k)}, y^{(k)}| h_k) = p(y^{(k)}|x^{(k)}, h_k)p(x^{(k)}| h_k)$. 
 The problem setting under which $p(y^{(k)}|x^{(k)}, h_k)$ is assumed invariant with respect to the choice of $h_k$ (i.e., $p(y^{(k)}|x^{(k)}, h_k)=p(y^{(k)}|x^{(k)})$) is often referred to as \textit{domain shift}. 
 We can also consider the problem in which only $p(x^{(k)}| h_k)$ is invariant with respect to $h_k$. 
For brevity, we refer to this problem-setting as \textit{function-shift}.
In general, meta learning problem is either the problem of \textit{function shift} or the problem of \textit{domain shift}, or both. 
For the problem of \textit{domain shift}, one would be required to make inference on the domain that is possibly outside the support of the observed dataset; 
this is essentially a problem of extrapolation, and it is an ill-posed problem unless we make some set of assumptions based on inductive bias, such as those related to metric.
Because we do not want to delve into the problem of \textit{which inductive bias to use} in our analysis, we focus on the problem of \textit{function shift} in this paper. 
Under the assumption of function shift, it can be shown that $p(h_t |x^{(t)}, \mathcal{D}_t) = p(h_t | \mathcal{D}_t)$.
That is, in the function-shift setting, the Bayes Risk Minimization problem we have formulated so far can be solved by seeking the encoder $p(h_t | \mathcal{D}_t)$ and the decoder $p(y^{(t)} | x^{(t)} , h_t)$.
Indeed, this objective coincides with that of Neural Process! 
We have just given the justification to the approach of Neural Process when the underlying model satisfies the function-shift condition.

\section{Smart Gaussian Approximation of the posterior}


Now that we have justified the learning of the encoder-decoder pair, the problem still remains as to which function family should be used for the approximation of the posterior and the likelihood distribution. 
In the conventional setting of supervised learning that uses predictive distribution (e.g. VAE), 
the encoder 
is a function of $x$ in the query domain only.
As we saw in, Eq.\eqref{eq:bayes_risk_base} however, 
the encoder in meta-learning is a function of not just one domain.
In particular, the encoder needs to accept a size-varying, unordered set $D_k$ from different domain in addition to $x$ from the query domain. 
Finding an appropriate family of function for encoder is therefore a nontrivial task,  and almost all methods developed to date take some measure to resolve this problem. 
Neural Process and GQN \cite{Eslami18GQN} \cite{Garnelo18CNP2} addressed the permutation-invariance problem by introducing the aggregation function. 
However, to the best of our knowledge, there has not been a study that have proposed an encoder that can represent a formally valid posterior distribution. 
For example, even when the appropriate set of conditions are met, the variance of the posterior distribution constructed in NP does not necessarily converge to $0$ as we increase the number of contextual information. 
To resolve this problem, we propose a novel design of the encoder that respects the rule of posterior distribution.

To do so, we introduce a little trick.
We begin from what is obvious from Bayes rule;
\begin{align}
  q(h | D) \propto  \left\{ \prod_{n=1}^{N}q(y_{n}|x_{n},h) \right\} q(h). \label{eq:task_posterior}
\end{align}
Now, suppose that we can partition each $D$ into equal-sized groups of size $M$.
That is, if $b_m = \{(x_{L(m-1)+i},y_{L(m-1)+i})|i=1,\cdots, L\}$, 
we assume that we can write $D$ as  $(b_1, \cdots, b_{M})$, $(N = LM)$. 
When this is the case, it holds that 
  \begin{align}
   & q(h | D) \nonumber \\
   \propto & \left\{  \prod_{m=1}^{M} \left\{ \prod_{i=1}^{L} q(y_{L(m-1)+i}|x_{L(m-1)+i},h) \right\} \right\}  q(h). \label{eq:task_posterior}
  \end{align} 
Now, if we write
 \begin{align}
   \prod_{i=1}^{L} q(y_{L(m-1)+i}|x_{L(m-1)+i},h)  \propto \frac{q(h | b_m)}{q(h)} \label{eq:batch posterior}
 \end{align}  
 and substitute the above into \eqref{eq:task_posterior}, we obtain
\begin{align}
  q(h |  D) \propto  \left\{ \prod_{m=1}^{M}q(h |b_m) \right\}/ q(h)^{M-1}.\label{eq:rewrited task_posterior}
\end{align} 
In order to make the computation of $q(h |  D)$  tractable, we will assume that both $q(h |b_m)$ and $q(h)$ are members of an exponential family.
As we will discuss later, when $m$ is large enough, we can use a variant of central limit theorem 
to validate this assumption. 
Then we can re-write the previous expressions as 
 \begin{align}
   q(h |b_m) & = Z(\eta(b_m))\exp( \eta(b_m)^T \xi(h))  \\  &\propto  \exp( \eta(b_m)^T \xi(h)), \nonumber \\ 
   q(h) & = Z(\eta_0)\exp( \eta_0^T \xi(h)) \label{eq:exponential family} \\ 
   &\propto  \exp( \eta_0^T \xi(h)), \nonumber
 \end{align} 
where $\eta(b_m)$ and $\xi(h)$ are repsectively the natural parameter and the natural statistic of the exponential family.
If we substitute this into Eq.\eqref{eq:rewrited task_posterior}, 
we obtain 
\begin{align}
   q(h |  D) = & Z\left( \eta_{M }\right) \exp( \eta_{M }^T\xi(h) )  \nonumber \\ 
  \propto &  \exp( \eta_{M }^T\xi(h) ),
\label{eq:rewrited task_posterior2}
\end{align}
where $\eta_{M } = \sum_{m=1}^{M}\eta(b_m) - (M-1)\eta_0$. 
If we chose the family of distributions for which the integral expression $Z(\eta)$ can be analytically computed, we can use the parameterized $\eta(b_m)$ to seek the member of the family that best approximates the true posterior distribution. 

In order to provide more intuition, we would like to describe a case in which $L=1$ and the exponential family of our shoice is Gaussian. 
By assuming that both $p_\theta(h | y_{n}^{(k)},  x_{n}^{(k)} )$ and $p_\theta(h)$ are Gaussians, we can represent $p_\theta(h | D)$ as a Gaussian distribution even when the likelihood term does not have a closed analytic form. 
To see this, let us suppose $p_\theta(h | x_{n}^{(k)}, y_{n}^{(k)}) = \mathcal{N}(h|f(x_{n}, y_{n}) ,G(x_{n}, y_{n}))$ and $p_\theta(h) = \mathcal{N}(h | f_0, G_0)$ where  the parameters of the functions $f$ and $G$ constitute the parameter vector $\theta$.
Let us also write $f_{n} = f(x_{n}, y_{n})$ and 
$G_{n} = G(x_{n}, y_{n})$ for short.  
Then we can analytically show $p_\theta(h | D) = \mathcal{N}(h | \mu(D), \Sigma(D))$ where 
\begin{align}
\mu(D) &= \Sigma (D) \left( \sum_{n=1}^{N} G_{n}^{-1} f_{n} -(N -1)G_0^{-1}f_0 \right), \label{eq:mu} \\
\Sigma(D) &= \left(\sum_{n=1}^{N}G_{n}^{-1} - (N-1)G_0^{-1} \right) ^{-1}. \label{eq:sigma}
\end{align}
If we parameterize the natural parameters $(f(x_{n}, y_{n}) ,G(x_{n}, y_{n}), f_0, G_0)$ by $\theta$,  we can seek the Gaussian distribution that best approximates the posterior distribution by optimizing the Bayes risk with respect to $\theta$.
Choosing Gaussian as the family of the posterior distribution not only makes the computation tractable, but also agrees with the general fact that $p(h | b_m)$ acts more like Gaussian distribution for large $m$. 
We will articulate this point further in the later discussion.

\section{Optimization of the parameter}

In the previous section, we have introduced a design of an encoder that respects all general properties of the posterior distribution conditioned against the unordered set $\mathcal{D}_k$ (eq.\ref{eq:rewrited task_posterior2}).
Now we can use ELBO to train the encoder and the decoder without any troubled conscience. 
Let us parameterize $\eta(b_m)$ in \eqref{eq:rewrited task_posterior2} by $\theta$, and use $p_\theta(h_k| \mathcal{D}_k)$ and $p_\tau (y^{(k)}| x^{(k)},h_k)$ to respectively represent the $\theta$-parametrized encoder and $\tau$-parametrized decoder.

In general, if we assume an infinite representation power for the parametric families $p_\tau (y^{(k)}| x^{(k)},h_k)$ and $p_\theta(h_k| \mathcal{D}_k)$,    the maximizer of
\begin{align}
\mathcal{L}_k(\theta, \eta) 
    :=& - \int p_\theta(h_k | \mathcal{D}_k) \sum_{n=1}^{N_k}   \left( \log q_\tau (y^{(k)} | x^{(k)}_n, h_k)
    \right) \nonumber \\
    & \left. + \log p_\theta(h_k) \right) d h_k - H(p_\theta (h_k | \mathcal{D}_k)),   
    \label{eq:ELBO}
\end{align}
in the asymptotic limit of $N_k \to \infty$ is given by $\tau^*$ and $\theta^*$
satisfying $p_{\tau^*}(y^{(k)}| x^{(k)},h_k) = p(y^{(k)}| x^{(k)},h_k)$ and $p_{\theta^*}(h_k| \mathcal{D}_k)= p(h_k| \mathcal{D}_k)$.
We therefore optimize \eqref{eq:ELBO}. 
The following table describes our algorithm based on a Gaussian distribution.
The table \ref{Alg:Training},\ref{Alg:Inference} summarize our algorithm.
Note that, in this expression, each $f_{n}$ is weighted by $G(x_n, y_n)^{-1}$, which intuitively represent the reliability of the point $(x_n, y_n)$ in the prediction of the query data.

\begin{algorithm}[t]
 \caption{Training algorithm}
 \begin{algorithmic}[1]
 \renewcommand{\algorithmicrequire}{\textbf{Input:}}
 \renewcommand{\algorithmicensure}{\textbf{Output:}}
 \REQUIRE initial parameters $\theta$ and $\tau$, pool of tasks $\mathcal{T}_{train} := \{ (k,\mathcal{D}_k) | k=1,\cdots, K\}$ and optimization algorithm \textit{OptAlg}
 \ENSURE  updated parameters $\theta$ and $\tau$
  \WHILE {Termination condition is unsatisfied}
  \STATE pick a task $k$ at random from $\mathcal{T}_{train}$.
  \FOR {$n=1, \cdots , N_k$}
  \STATE Compute $f_{nk}=f_{\theta}(x^{(k)}_n, y^{(k)}_n)$ and $G_{nk}=G_{\theta}(x^{(k)}_n, y^{(k)}_n)$ by the current encoder
  \ENDFOR
  \STATE Aggregate $\{(f_{nk}, G_{nk}) | n=1,\cdots, N_k\}$ and Compute $\mu_k$ and $\Sigma_k$ according to Eqs.\eqref{eq:mu} and \eqref{eq:sigma}
  \STATE Sample $h_k$ from $p(h_k | \mathcal{D}_k) = N(h_k |\mu_k, \Sigma_k )$ and compute the stochastic gradient of Eq.\eqref{eq:ELBO} w.r.t $\theta$ and $\tau$
  \STATE Update the parameters $\theta$ and $\tau$ with the stochastic gradient by using a optimization algorithm \textit{OptAlg}
  \ENDWHILE
 \RETURN $\theta$ and $\tau$
 \end{algorithmic} 
 \label{Alg:Training}
 \end{algorithm}
 
 \begin{algorithm}[t]
 \caption{Inference algorithm}
 \begin{algorithmic}[1]
 \renewcommand{\algorithmicrequire}{\textbf{Input:}}
 \renewcommand{\algorithmicensure}{\textbf{Output:}}
 \REQUIRE parameters $\theta$ and $\tau$, dataset of query task $t$, 
 $D^{(t)} = \{(x^{(t)}_1, y^{(t)}_1), \cdots, (x^{(t)}_{N_t}, y^{(t)}_{N_t}) \}$ and query input $x^{(t)}_*$
 \ENSURE  Estimate of the output $y^{(t)}_*$ corresponding to the input $x^{(t)}_*$
  \FOR {$n=1, \cdots , N_t$}
  \STATE Compute $f_{nt}=f_{\theta}(x^{(t)}_n, y^{(t)}_n)$ and $G_{nt}=G_{\theta}(x^{(t)}_n, y^{(t)}_n)$ by the current encoder
  \ENDFOR
  \STATE Aggregate $\{(f_{nt}, G_{nt}) | n=1,\cdots, N_t\}$ and Compute $\mu_t$ and $\Sigma_t$ according to Eqs.\eqref{eq:mu} and \eqref{eq:sigma}
  \STATE Sample $h_t$ from $p_{\theta}(h_t | \mathcal{D}_t) = N(h_t |\mu_t, \Sigma_t)$
 \RETURN ${\hat{y}}^{(t)}_* = \int y q_{\tau}(y|x^{(t)}_*, h_t)dy$
 \end{algorithmic} 
 \label{Alg:Inference}
 \end{algorithm}

\section{Theoretical properties on Bayes optimal estimator and Gaussian posterior approximation}
In this section, we will discuss the property of the predictive distribution.
In particular, we discuss the rate at which $p(y |x, D)$ produced from $p(h|D) $ approaches $p(y | x, h^*)$ where $h^*$ is the true latent variable.

When the problem is of function-shift type, the Bayes risk looks like 
\begin{align}
 & E\left[KL[p(y|x,h)|p(y|x,D)] |h \right] \nonumber \\ 
=& \iiint p(h|D)p(x,y|h)\log \frac{p(y|x,h)}{p(y|x,D)}dydxdD
\end{align}
where $p(y|x,D) =  \int p(y|x,h')p(h'|D)dh'$.
Then, in general, the Bayes risk of this optimal estimator asymptotically decay with the following rate with respect to the number of samples.
\begin{theorem}[Asymptotic Bayes risk of Bayes optimal estimator] \label{thm:AsymptoticBayesRisk}
Suppose that $\{D_N\}$ is a monotonic sequence of a set of i.i.d. samples from $p(\cdot, \cdot |h^*)$ with $|D_N| = N$. Then the MLE    
\begin{align}
    {\hat h}_N =  \argmax_{\tilde h}  \log p(D_N|\tilde h) 
\end{align}
converges to $h^*$ almost surely in the limit of $N \to \infty$, and 
\begin{align} 
 & E\left[KL[p(x,y|h^*)|p(x,y|D_N)]|h^* \right] 
 = \frac{d}{2N} + O\left(\frac{1}{N^2}\right).
\end{align} \label{eq:AsymptoticBayesRisk}
\end{theorem} 
See \citet{Strasser77BayesRisk, Hartigan98AsymptoticBayes, Komaki15AsymptoticBayes} for the rigorous regularity conditions required for this result.
Note that the RHS of the expression above is an expectation conditioned against the query task variable, $h^*$.
Interestingly, this order is the same as the decay order of $E\left[KL[p(x,y|h^*)|p(x,y|\hat h(D))]|h \right]$ when $\hat h(D)$
is a maximum likelihood estimate of $h^*$ given $D$. 
\footnote{Optimal Bayes estimator differs from the maximum likelihood estimate, but their asymptotic behavior only differs with a term of order $O(N^{-2})$.}
Moreover, this asymptotic order is independent of the choice of $h^*$.
In other words, the rate of asymptotic improvement with respect to $N$ does not depend on the choice of the query task if all $D$ are to be sampled from $p(x,y|h^*)$ in i.i.d manner. 

Now, note that the part of the the optimal predictive distribution that depends on $N$ is the posterior distribution $p(h|D)$ only. 
Let us therefore look closer into the asymptotic behavior of $p(h|D)$.
Under some regularity assumption, it is known that the posterior converges to a Gaussian distribution \cite{Vaart00Asymptotics}. 
For notational simplicity, let $z=(x,y)$ and 
$\Delta_{n,h^*}=\frac{1}{n}\sum_{i=1}^n I_{h^*}^{-1}
\frac{\partial \log p(z|h)}{\partial h}\mid _{h=h^*} $
where $I_{h^*}$ is a Fisher information matrix $I_{h^*} = \int p(z_n|h^*)
\left(\frac{\partial \log p(z|h)}{\partial h}\mid _{h=h^*}\right) \left(\frac{\partial \log p(z|h)}{\partial h}\mid _{h=h^*}\right)^Tdz$.
\begin{theorem}[Bernstein-von Mises] \label{thm:Bernstein-von Mises}
Let the domain of $z$ be $\mathcal{Z}$, and let distribution $p(z|h)$ be differentiable in quadratic mean at true parameter $h^*$ with nonsingular Fisher information matrix $I_{h^*}$. 
Suppose that for every $\epsilon > 0$ there exist a sequence of tests 
$\phi_N: \mathcal{Z}^N \to [0,1]$ such that
\begin{align} 
&E_{p(\cdot | h^*)} [\phi_N(D_N)] \to 0,  \nonumber \\
&\sup_{\| h-  h^* \| \geq \epsilon}  E_{p(\cdot | h^*)} [1- \phi_N(D_N)] \to 0  \nonumber
\end{align} 
Furthermore, let the prior measure be absolutely continuous in a neighborhood of $h^*$ with a continuous positive density at $h^*$. Then the corresponding posterior distributions satisfy
\begin{align} 
\lVert P_{\sqrt{N}(h-h^*)|z_1,\cdots, z_n} - 
\mathcal{N}(\Delta_{n,h^*},I_{h^*}^{-1}) \rVert \xrightarrow{P(\cdot|h^*)} 0.
\end{align} \label{eq:AsymptoticNormal}
\end{theorem} 
where $P_{h|x_1,\cdots,x_n} = \frac{(\prod_{i=1}^n p(x_n|h))p(h)}{\int (\prod_{i=1}^n p(x_n|h))p(h) dh}$ and 
$\xrightarrow{P(\cdot|h^*)}$ indicates the converence in probability  $P(\cdot|h^*)$.
Moreover, it is known that, if the MLE $\hat h(D_N)$ is a consistent estimator of $h^*$, then 
$$\left\lVert P_{h|z_1,\cdots, z_n} - 
\mathcal{N}\left(h^* ,\frac{1}{N} I_{h^*}^{-1}\right) \right\rVert \xrightarrow{P(\cdot|h^*)} 0 $$
as well. 
In simpler words, this result states that the rescaled and centered posterior distribution converges in probability a Gaussian distribution with center $h^*$.

Thus, if the batch size $L$ in \eqref{eq:batch posterior} is large enough and if the system is regular enough,  we may approximate $p(h|b_m)$ 
rightfully as a Gaussian distribution. 
When we approximate the posterior distribution by Gaussian, the expression \eqref{eq:sigma} suggests that the variance of our estimated posterior converges to $0$ with the same rate as MLE ($O(1/N)$) if $|G_{nk}|=|G(b_{nk})| > |G_0|$.
The requirement $|G_{nk}||=G(b_{nk})| > |G_0|$ is natural because this relation just means that the variance of the posterior distribution  shall decrease with respect to the number of parameters. 
It is not too difficult to train a model that satisfies this requirement. 
For example, we may construct the model so that 
$G=G_0 + \Phi(b_{nk})\Phi(b_{nk})$ always hold.
In the next section, we compare our Gaussian posterior approximation with other Gaussian approximations that were used in previous studies.

\section{Relationship with other methods} \label{sec:relation}
To the best of author's knowledge, there are no studies to date that use a batch in the way of \eqref{eq:batch posterior} to compute the posterior distribution of latent variable.  
For the sake of fair comparison, we therefore discuss the case of $L=1$ in this section. 
Assume $p(h) = \mathcal{N}(0,I)$.
\subsection{Linear Gaussian Model (LGM)}
Linear Gaussian model is a classic model that computes the posterior distribution \eqref{eq:task_posterior} analytically by assuming a linear model for the likelihood and a Gaussian distribution for the prior distribution. 
Formerly, Linear Gaussian model assumes the following;
\begin{align} 
p(y_{n}|x_{n},h) &= \mathcal{N}(y_n| W(x_n)h + b(x_n), G(x_n)) \nonumber \\
p(h) &= \mathcal{N}(0,I)
\end{align}
In this case, the posterior distribution can be written as $p_\theta(h | D) = \mathcal{N}(h | \mu_{LGM}(D), \Sigma_{LGM}(D))$ where
\begin{align}
\mu_{LGM} (D) &= \Sigma (D)  \sum_{n=1}^{N} W(x_n)(y_n - b(x_n)) \label{eq:linear_mu} \\
\Sigma_{LGM}(D) &= \left( I + \sum_{n=1}^{N} W(x_n)^{T}G(x_n)^{-1}W(x_n) \right)^{-1}  \label{eq:linear_sigma}
\end{align}
As is clear in the expression \eqref{eq:linear_sigma},  the variance of the posterior distribution decays with order $O(1/N)$ for the Linear Gaussian model as well.
Meanwhile, in Linear Gaussian model, the posterior mean and \eqref{eq:mu} is a linear function with respect to $y$, and the posterior variance  \eqref{eq:sigma} is a variance that is independent of $y$.
Because the posterior mean and the posterior variance of our method are both non-linear functions of $y$, the space of models that can be represented by our model is much greater than the one considered by LGM.  
In fact, we can use our model to realize the posterior of linear Gaussian model by choosing $(W(x_n)^{T}G(x_n)^{-1}W(x_n) + I)W(x_n)(y_n - b(x_n))$ for $f(x_n,y_n)$ and choosing $W(x_n)^{T}G(x_n)^{-1}W(x_n) + I)^{-1}, 0,I)$ for $G(x_n,y_n)$.


\subsection{Generative Query Network (GQN)}
Generative Query Network (GQN) \cite{Eslami18GQN} is a celebrated method that succeeded in solving the complex task of rendering the scene from an unseen direction based on a few arbitrary pairs of camera-location an captured scene. 
In their paper, \cite{Eslami18GQN} demonstrated GQN's ability to carry out this task in environments with varying colors of wall as well as the types and the locations of objects in the system. 
From now on, we will refer to the task solved in \cite{Eslami18GQN} as Neural Scene rendering task.
GQN and our method are similar in that they too take the approach of encoding the observations of the given environment into a latent variable.
To make prediction for the query input (i.e new location of camera), GQN   conditions the decoder function against the latent variable. 
When we interpret GQN in our framework, their latent variable $r$ corresponds to $h$ in our method.
\footnote{Although GQN also uses other latent variable $z$, we did not mention $z$ in our discussion here because their $z$ does not depend on the input.} 
If we regard their deterministic output as a sample from a posterior distribution, we may say that GQN is using 
$p(h | D) = \delta(h - \mu_{GQN}(D))$
as their posterior distribution, where $\delta$ is the dirac delta and $\mu_{GQN}$ is given by
\begin{align}
\mu_{GQN} (D) &= \sum_{n=1}^{N} f(x_n,y_n). \label{eq:GQN_mu} 
\end{align}
Unlike Linear Gaussian Model, GQN thus uses a nonlinear function to construct the latent code $h$.  
At the same time, GQN model does not explicitly formulate the uncertainty of the encoded $h$.
Also, by its design, the model will definitely diverge as we increase the number $N$ of the query dataset sampled from $p(\cdot, \cdot | h^*)$.
We would discuss this problematic behavior further in the experimental section.

\subsection{Neural Process (NP)}
The family of Neural Process \cite{Garnelo18CNP2} is closely related to our work, and some of its variants have been particularly successful in computer-vision applications.
In particular, \citet{Kim19ANP, Louizos19FNP} devised ways to encode inter-pixel correlations to greatly improve the model's performance on the image-completion task.
Also, \citet{Gordon19ConvolutionalCNP} discovered a general way to construct a shift-equivariant $+$ permutation invariant encoder and leveraged its ability to complete a large image using the training set consisting of small images. 
Again, if we interpret NPs in  our context, we may say that these methods use the posterior distribution of the form 
$p_\theta(h | D) = \mathcal{N}(h | \mu_{NP}(D), \Sigma_{NP}(D))$ where\footnote{They too also propose a deterministic encoder like GQN, but we omitted their determinisic formulations because they can be realized by taking $N$ to $\infty$. }
\begin{align}
\mu_{NP}(D) &= f\left( 
\frac{1}{N}\sum_{n=1}^{N} \phi (x_n,y_n) \right), \label{eq:CNP_mu} \\
\Sigma_{NP}(D) &= g\left(\frac{1}{N}\sum_{n=1}^{N} \phi (x_n,y_n)  \right). \label{eq:CNP_sigma}
\end{align}
These methods too use non-linear functions about both $x$ and $y$ in order to construct the latent variable $h$.
Also, unlike our method, their methods apply a post-linear transformation after the aggregation of $\phi$s. 
While their formulation seems similar to our method if we disregard the post-linear transformations $f$ and $g$, their formulation differs from our method most greatly in that it does not have the mechanism to weigh the $(x_n, y_n)$ by its importance.
Thus, if $N$ is small and if some observation can be much less reliable than others, this formulation might fail to make a good prediction. 
Meanwhile, our method generally assigns heavy weight to $(x_n, y_n)$ with small uncertainty $G^{-1}(x_n, y_n)$, and vice versa.
Also, because of the post-linear transformation, the variance \eqref{eq:CNP_sigma} does not decay with order $1/N$ for an arbitrary choice of $g$.
The only guarantee that one can make to \eqref{eq:CNP_sigma} is that it will converge to some number with the Law of large numbers if $(x_n,y_n)$ are sampled in the i.i.d manner. 

\section{Experiments}
We conducted a series of experiments in order to study the following: 
\begin{enumerate}
    \item The representation power of our model relative to Linear Gaussian model 
    \item The effect of using a larger batch size in equation \eqref{eq:batch posterior}
    \item Competitiveness of our model on a regression task and the neural scene rendering task 
\end{enumerate}
Also, in order to study the basic properties of our model on generic dataset, we did not compare our model against the models that are specialized for specific dataset (i.e those that use a specific mechanism to model spatial correlation /invariance in the dataset)
For the architectures of the models we used in our comparative study, please see the supplementary material.

\subsection{Linear function with discontinuity points}
We conducted a few-shot regression task for 1D functions with multiple discontinuity points.
This task is deceivingly difficult because the discontinuity points differs across the tasks(functions). 


\begin{figure}[ht]
\centering
\includegraphics[width=80mm]{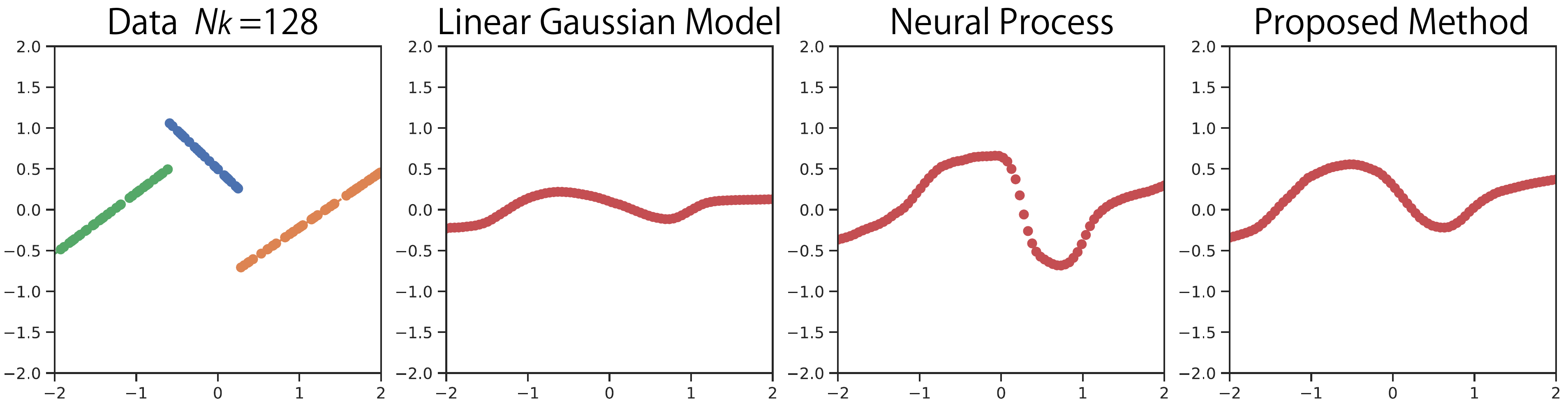}
\caption{Visualization of the results on the regression task for 1D linear functions with multiple discontinuities.
} 
\label{fig:visualization_1D_piecewise_linear_16}
\end{figure}
In this task, the posterior distribution tends to be complex and LGM performs poorly in comparison to our model and Neural process, 
This is likely because of the lack of the representation power of LGM's encoder.
The figure \ref{fig:result_1Dpiecewise} plots the performance of various model for this few-shot function learning task. 
As we can see in the figure, both NP and our model consistently outperform LGM.

\begin{table}[ht]
    \centering
    \small
    \begin{tabular}{c||c|c|c}
         $N_k$
         & LGM
         & NP & Ours \\ \hline 
         8
         & -1.127 ($\pm$0.327) & -1.027 ($\pm$0.325)& -1.090 
          	 ($\pm$0.351 ) \\ \hline 
         16
         & -1.117 ($\pm$0.304) & -0.979 ($\pm$0.216) & -1.050 ($\pm$0.361) \\  \hline
        32
         & -1.085 ($\pm$0.309) & -0.967 ($\pm$0.195) & -0.998  ($\pm$0.195) \\ \hline
         64
         & -1.087 ($\pm$0.295) & -0.978 ($\pm$0.243) & -0.972 ($\pm$0.134) \\ \hline 
        \end{tabular}
    \caption{Performance of trained policies on and unknown Jam environments}
    \label{JAM2}
\end{table}

\subsection{Neural scene rendering task} 
As described in the previous section, the goal of the neural rendering task is to train a model that can predict the scene from an unseen (random) direction in a completely new environment.
In this experiment, we prepare datasets consisting of numerous (scene, camera location+direction) pairs collected from different artificial rooms that are constructed with different wallpapers and different set of randomly colored geometrical objects.
The locations of geometrical objects differ across rooms. 
For each room $q$, random set of camera location+direction pairs are used to construct the contextual dataset $D_q$.
The formal goal of this task to learn a model that can use $D_{q^*}$ of previously unseen room to predict the scene $y^*$ from the query camera location+direction pair $x^*$.
For the model architecture used in this experiment, please see the supplementary material for the details.
The Figs.\ref{fig:field} plots the MSE of various methods against the number of observations $D_{q^*}$.
As we described earlier, the prediction of GQN diverges as we increase the number of observations.
Also, as we can see in the plot, our method performs better than NP when the number of observations is small.
This is possibly because NP lacks the weighting mechanism that we mentioned earlier.
Unlike NP, the variance of our prediction also approaches $0$ as we increase the number of observations.
\begin{figure}[ht]
\centering
\includegraphics[width=78mm]{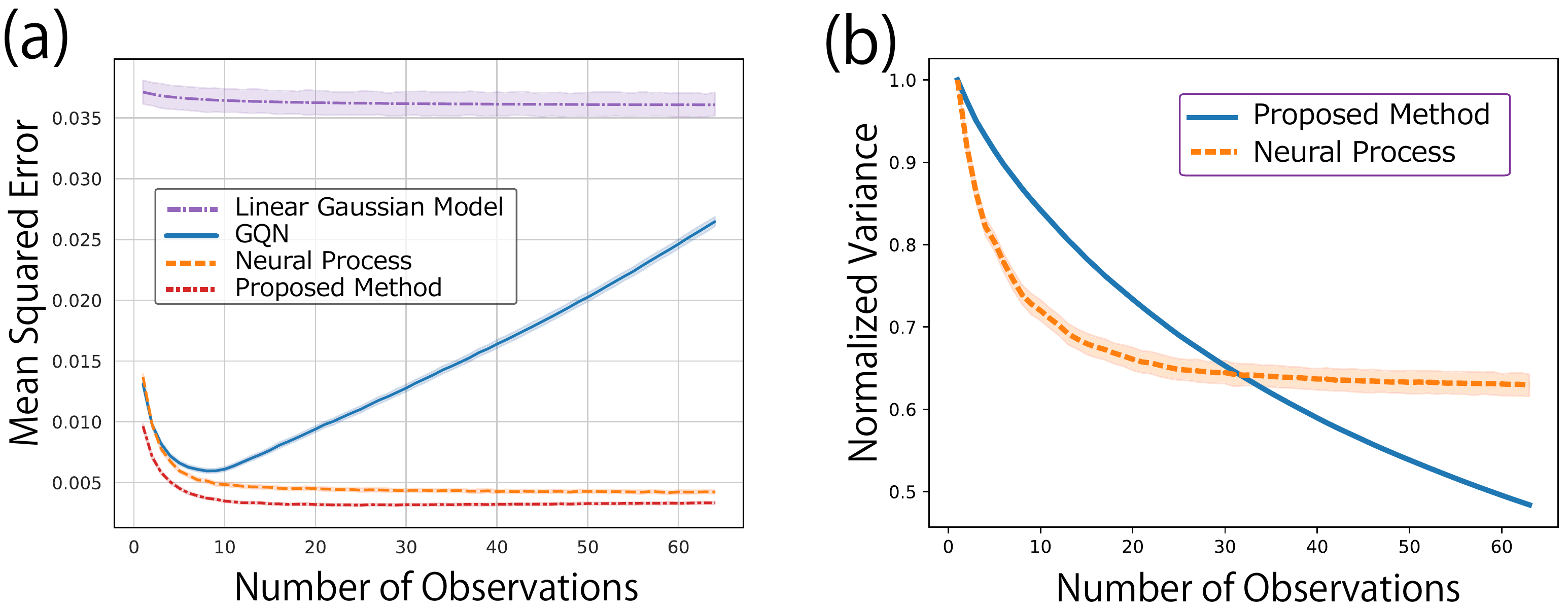}
\caption{(a) MSE plotted against the number of observations $N$ in the query task. We see that our method not only achieves consistently better MSE than the other two methods for all $N$, the MSE of our method decreases monotonically with respect to $N$. (b) The size of the variance of the latent variable $h$ plotted against the number of observations after scaling so that the variance of all models coincide when $N=1$. } 
\label{fig:field}
\end{figure}

\begin{figure}[ht]
\centering
\includegraphics[width=78mm]{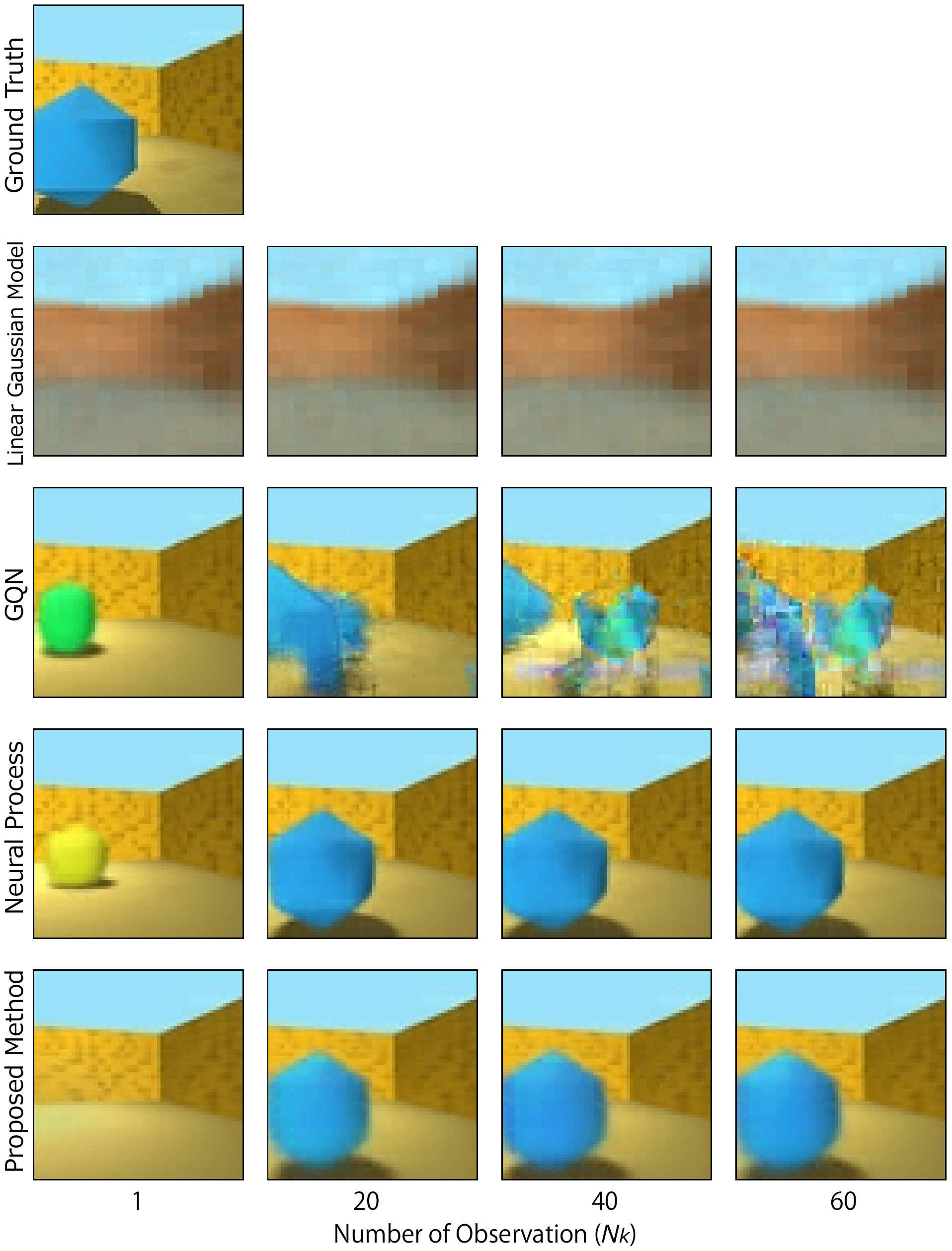}

\caption{Example of neural renderings produced by various methods.  As we see in the figure, the rendering by GQN collapses when the number of observations is large. } 
\label{fig:field}
\end{figure}

\section{Conclusion}
In this research, we used Bayes Risk Minimization to re-formalize the meta-learning problem.
The BRM-based perspective justifies the use of the predictive distribution in meta learning problems, and justifies the approach of previous methods like Neural Process and GQN \cite{Yoon18BayesianMeta, Kim19ANP}.
We also formerly described how the Bayes risk decreases with the number of observations obtained for the query task. 
Our study might provide some insight into the order of tasks to solve in Curriculum learning and Life long learning, as well as the appropriate size of task-dataset to use at each step.  
We also presented a novel method for approximating the posterior distribution. 
By choosing the appropriate exponentially family to represent the posterior distribution, we computed the Bayes-optimal natural parameter in an analytically computable form.
The family of the posterior distribution we propose in this study is also capable of representing a large family of distributions that includes the one used in classical Linear Gaussian model. 
The result of the our 1D function regression experiment suggests the superiority of the representation power of our model over that of LGM.
While seemingly similar to those used in GQN and NP, our posterior approximation is unique in that it can not only allow the model to evaluate a theoretically meaningful measure of uncertainty, it also allows the model to weigh each observation by its reliability in prediction.
Also, because our method is faithful to the theoretical properties of the posterior distribution, the variance of our posterior distribution decays with the same rate of $O(1/N)$ as the Bayes-optimal posterior distribution.
The stable performance of our method suggests that there is much room left for the study of the meta-learning models that observes the classical theoretical results of statistics.

\bibliography{ref}
\bibliographystyle{icml2020}




\clearpage

\input{Appendix_arXiv}

\end{document}

%% file: Appendix_arXiv.tex
\section{Appendix}

\subsection{Experiment details}
\begin{figure}[ht]
\centering
\includegraphics[width=78mm]{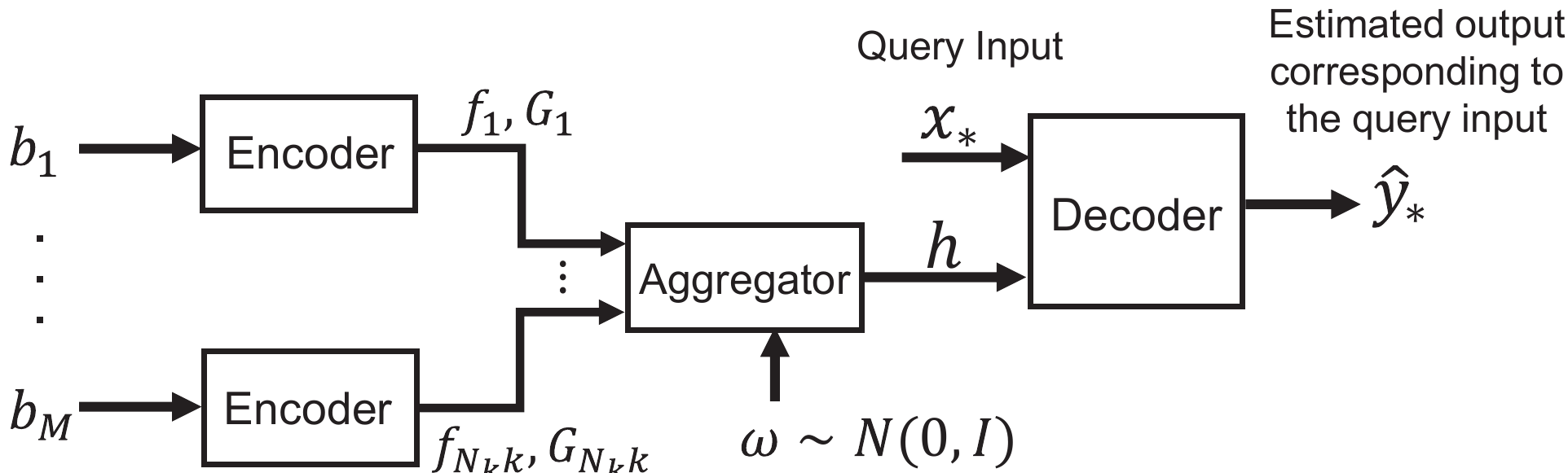}
\caption{Overall schematic of our model for both experiment.}
\label{fig:overall}
\end{figure}

In this Appendix section we will present the details of the experiments. 
The Figure \ref{fig:overall} is a general schematic of our model used for both 1D function regression task and Neural Scene rendering. 
In what follows, we will present more details of the models and the experimental settings for both sets of experiments.

\subsubsection{1D function} 
\begin{figure}[ht]
\centering
\includegraphics[width=78mm]{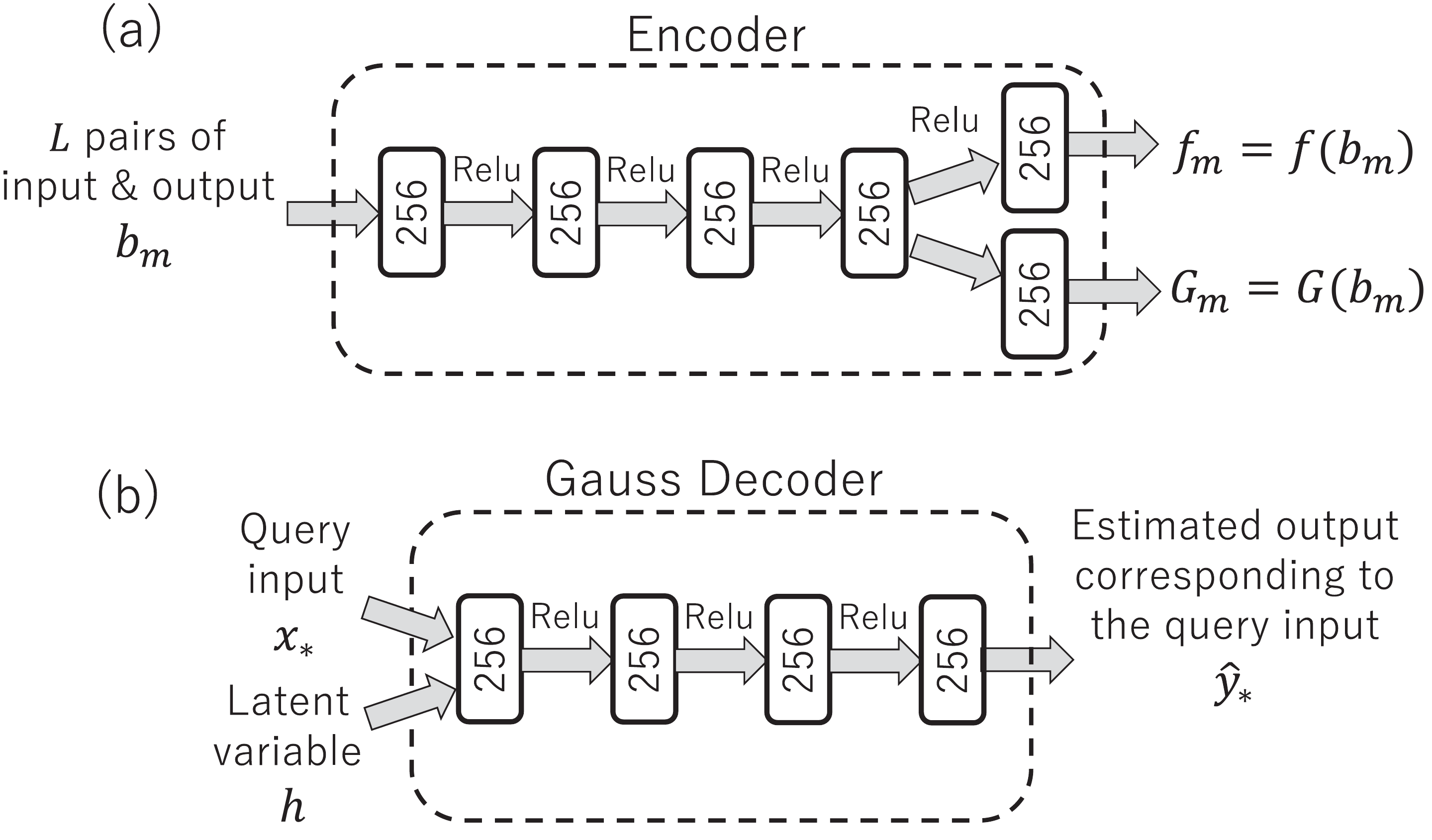}
\caption{Encoder and decoder design for 1D function estimation. For the flow decoder, we used continuou
s normalizing flow (CNF). Panel (c) is the design of $\frac{dy}{dt}$ in our CNF. }
\label{fig:1Dfxn}
\end{figure}

\begin{figure*}[ht]
\centering
\includegraphics[width=150mm]{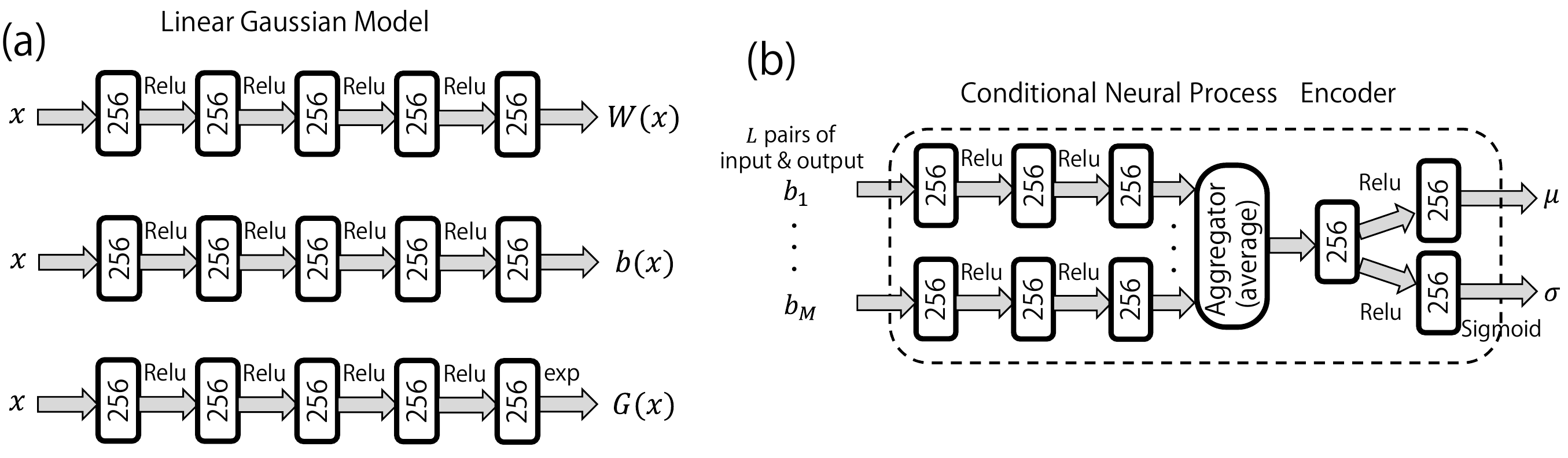}
\caption{Overall schematic of LGM and CNP.} 
\label{fig:LGMCNP}
\end{figure*}

\textbf{Architecture}\\
See Figure \ref{fig:1Dfxn} for the detailed schematic of our model for the 1D regression experiment.
As for the models we used in comparative studies, we used the architecture represented in Figure \ref{fig:LGMCNP} (a) for the LGM, and used the architecture in Figure \ref{fig:LGMCNP} (b) for the CNP.

\textbf{Optimization}
For the optimization, we used Adam with the fixed learning rate of 0.00005 and trained all models with batchsize 64 (64 tasks). 
More particularly, for the construction of each batch we chose a random integer value $k$ from the range $3 \sim 50$ and took $k$ samples from each one of $64$ tasks.
For the optimization of ELBO about encoder, we used a version of \textit{re-parametrization trick} to enable easy back propagation. 
More particularly, we produced the Gaussian posterior distribution by transforming the Gaussian distribution with deterministic function ($\omega \sim N(0, I)$ in Fig \ref{fig:overall}.)



\subsubsection{Neural Scene Rendering} 
\textbf{Data generation}  For the basic dataset in this set of experiment, we used the \textit{rooms\_free\_camera\_no\_object\_rotations} dataset  published in \cite{Eslami18GQN}.  
Each instance of observation this data consists of (1) location of the camera, (2) direction of the camera and (3) the corresponding scene.
For the train/test split,  we followed the same procedure as the one used in \cite{Eslami18GQN}; we trained the model with 10,800,000 scenes, and tested the model with 1,200,000 scenes. 
There are 10 per each room in the dataset.
At the time of the training, we selected the context size randomly from $1\sim 10$ and chose $1$ observation as query. 

We also constructed our own dataset using OpenGL so that we can increase the number of contexts. 
We trained all models on our hand-made dataset in the same way we trained the models on the deepmind dataset.

\textbf{Architecture and hyperparameter}
\begin{figure*}[ht]
\centering
\includegraphics[width=120mm]{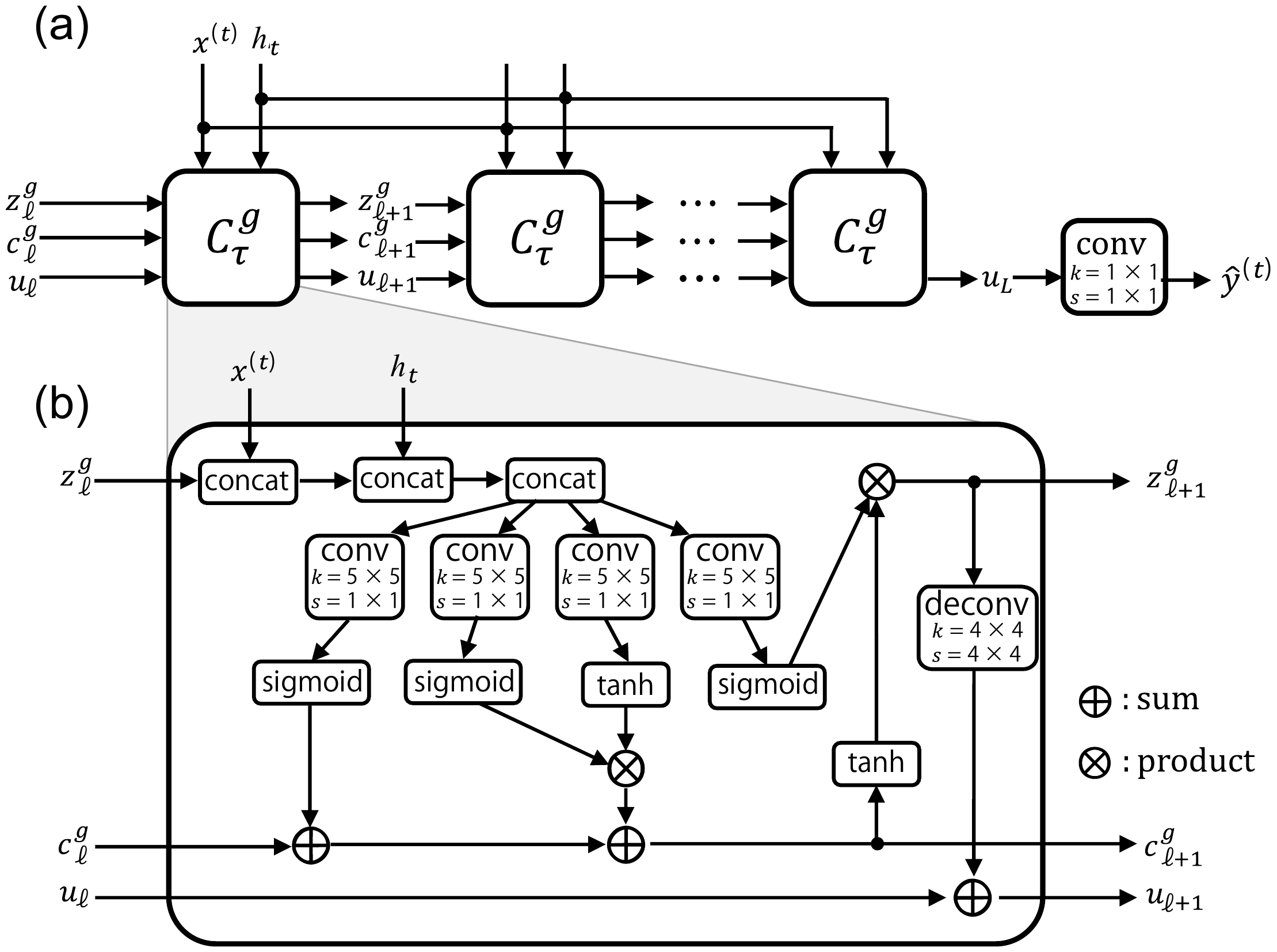}
\caption{Architecture of the decoder for the neural scene rendering task. The architecture very similar to the Generation network, and $c_\tau$ are the computation cores that takes in (1) the output $z_\ell^g$ of the LSTM network from the previous core, (2) the state $c_\ell^g$, and (3) the skip connection pathway variable $u_\ell$. 
We are using $x^{(t)}$ to denote a scene for the query room(task) $t$, and $h_t$ to denote the latent vector corresponding to $t$. 
Unlike the original architecture, however, we do not concatenate previous layer's output to $z_\ell$.  
Instead, $x^{(t)}, h_t$ concatenated to $z_\ell$ at all layer $\ell$.
} 
\label{fig:decoder architecture}
\end{figure*}

Figure \ref{fig:overall} is a brief schematic of the model used for our encoder and decoder.

We based our encoder design on the Representation Network of the original GQN. 
More particularly, we constructed our encoder by adding one Convolution layer(5 x 5 x 72) to the Representation Network of type \textit{Tower}, and partitioned its 512 dimensional output evenly to produce $f$ and $g$ in our formulation. 
We based our decoder design on the Generation Network of the original GQN (Fig \ref{fig:decoder architecture})
Recall that, in our framework, the encoder design corresponds to the posterior design, and the decoder design corresponds to the likelihood design.
Just as in the original GQN, we used 12 LSTMs with different model parameters.
Meanwhile, we made a slight modification to the encoder to to observe the fact that treatment of the latent vector $h$ in our model is slightly different from that of the original GQN.
The original GQN produces a sequence of latent variable $h_t$ recursively through the stack of LSTMs, and concatenate $h_t$s to produce a single latent variable $h$ to condition the output of the final convolutional LSTM layer.
Because our model does not produce $h$ in such a recursive manner, we passed the same $h$ produced by the encoder to all LSTMs in our encoder. 
See Fig \ref{fig:decoder architecture} for the detail of our decoder design.

\textbf{Optimization}
For the training of our model, we followed the same procedure as in \cite{Eslami18GQN}, and conducted Adam with standard parameters and annealed the learning rate from $0.0001$ to $0.00005$ over 2million steps.
We used batches of size 64.